\documentclass[a4paper]{article}                                    % Needed to meet printer requirements.

\usepackage{a4wide}

\usepackage{hyperref}       % hyperlinks
\usepackage{url}            % simple URL typesetting
\usepackage{booktabs}       % professional-quality tables
\usepackage{amsfonts}       % blackboard math symbols
\usepackage{nicefrac}       % compact symbols for 1/2, etc.
\usepackage{microtype}      % microtypography
\usepackage{graphicx}
\usepackage{amssymb}
\usepackage{amsfonts}
\usepackage[cmex10]{amsmath}

\title{Learning What's Above and What's Below:\\ Horizon Approach to Monocular Obstacle Detection}

\author{
  G.C.H.E. de Croon and C. De Wagter
	\thanks{Micro Air Vehicle laboratory, Control and Simulation, Faculty of Aerospace Egineering, Delft University of Technology, the Netherlands, \texttt{g.c.h.e.decroon@tudelft.nl}
	}
}

\begin{document}

\maketitle

\begin{abstract}
A novel approach is proposed for monocular obstacle detection, which relies on self-supervised learning to discriminate everything above the horizon line from everything below. Obstacles on the path of a robot that keeps moving at the same height, will appear both above \emph{and} under the horizon line. This implies that classifying obstacle pixels will be inherently uncertain. Hence, in the proposed approach the classifier's uncertainty is used for obstacle detection. The (preliminary) results show that this approach can indeed work in different environments. On the well-known KITTI data set, the self-supervised learning scheme clearly segments the road and sky, while application to a flying data set leads to the segmentation of the flight arena's floor.
\end{abstract}

%\keywords{Self-supervised learning, visual sonar, horizon line, sky versus ground classification} 

\section{Introduction}
Despite decades of intensive research, robust autonomous obstacle detection by mobile robots is still a daunting challenge. Currently, the most successful solutions rely on high-end sensors such as laser scanners \cite{ACHTELIK2009}, RGBD sensors \cite{SHEN2011}, or stereo vision systems \cite{SHEN2013}. Although these solutions are quite successful, they can take up more resources than desirable. For instance, for small, light-weight flying robots a solution based on a single camera is much more desirable. 

For this reason, monocular vision for obstacle detection is heavily investigated. The majority of such work focuses on the exploitation of motion cues for the extraction of relevant optical flow observables such as time-to-contact \cite{CAMUS1995,GREEN2008}, visual odometry \cite{FORSTER2014} or visual SLAM \cite{engel2014lsd}. Early work on these topics most often made use of a traditional feature-detection-and-tracking approach (e.g., \cite{STRASDAT2010}), which resulted in very sparse point-like observations for obstacle detection. More recently, direct approaches are being studied (as in \cite{engel2014lsd}), which obtain both the camera's egomotion and the world structure by directly matching the illuminance values in subsequent images. These matches are most reliable in textured areas, making the so-created maps look like line-drawings. This implies a denser perception of the environment than with point features, but still textureless areas are not covered.

The use of motion cues is attractive, because they generalize to any unknown environment. However, they also have their limitations, such as the requirement of sufficient texture. A natural complement to motion cues would be the use of \emph{visual appearance} cues, i.e., the recognition of colors, textures, and patterns. These cues are known to play an essential role in human perception.

% Motion cues will be important to any monocular mobile robot for detecting obstacles, as they generalize to any unknown environment. However, these cues also have their limitations, such as the requirement of sufficient texture. Hence, it is worth investigating additional monocular cues such as the visual appearance of objects in the environment. Such cues play an important role in human

A minority of work on monocular obstacle detection has focused on using visual appearance cues. This type of work does come in many different flavors, four of which will be discussed here. First, some approaches directly map appearance features to actions, as was done with imitation learning in \cite{ROSS2013} and is now increasingly being done with end-to-end deep learning \cite{levine2016end}. This type of approach is definitely very suitable if the training environment is similar in some way to the test environment. \cite{sadeghi2016cad} provided perhaps the most impressive example of this approach, using a simulator of office-like indoor environments with strongly varying textures in order to have the learned deep net also work in a real office environment. However, applying the learned network to a completely different environment would likely result in a lower performance. Moreover, the network trained in simulation will in any case not exploit the specific appearance of the real-world office environment, as learning only takes place in simulation. 

A second approach relates image appearance to distances (e.g., \cite{bipin2015autonomous,dey2016vision}). An early study in this field used a data set obtained with a laser scanner and an aligned camera in order to learn a mapping from feature vectors (consisting of Laws masks and oriented edge filters) to a distance per pixel \cite{saxena2006learning}. The best results on this type of learning have been obtained by using deep learning \cite{eigen2014depth}. Of course, learning depths on a training data set does not guarantee that the learned method will work in the actual test environment. For instance, the absolute scale of the distance estimates can be far off if the training set had outside images while the test set consists of indoor images. In that respect, it is interesting to note that if the robot has an additional depth sensor or stereo camera on board, the learning of distances can be done on board in the test environment by means of self-supervised learning \cite{van2016self,garg2016unsupervised,godard2017unsupervised}. % A preliminary investigation into this self-supervised learning of distances is performed in \cite{HECKE2016}, in which texton distributions are mapped to a single average depth value per image to allow for a rudimentary navigation strategy.  

A third approach, which actually predates the previous two discussed approaches, is to assume the presence of a local ground plane. If such a plane is present and well-segmented in the image, the distances and directions to ground obstacles can be determined by using the image coordinates at which the obstacles touch the plane. In toy-like environments, as in the robocup competitions \cite{lenser2003visual,fasola2005fast}, a color can be pre-selected by a human, leading to a computationally efficient and accurate segmentation. However, if one wants to use this approach in a priori unknown environments, a learning component is essential. \cite{lorigo1997visually} learned the appearance of the floor by assuming that the floor directly in front of the robot was free of obstacles. \cite{ulrich2000appearance} argued that this was a strong assumption, and showed that it was important to also learn floor appearance further away, as the floor often has quite a different appearance higher up in the image. They solved this problem in an elegant way, by using the fact that successfully driving over the floor was a proof that it was free of obstacles. Hence, the robot could use its wheel odometry to retrieve the relevant regions from past images - learning also floor appearance further away. A drawback of this approach is that it will prevent the robot to drive on a floor of a different color than experienced before, limiting the robot's movements.

A fourth approach also segments the image, but now to make the difference between `sky' and `non-sky'. This idea, suggested in \cite{MCGEE2005}, was first elaborated in \cite{DECROON2011}. The idea is based on the fact that non-sky regions above the horizon line are obstacles to flying robots. \cite{DECROON2011} learned a decision tree based on a wide range of computer vision features for classifying pixels as sky or non-sky. This sky segmentation approach to obstacle detection allowed for very timely detection of far-away obstacles. This led to calm and timely obstacle avoidance actions. The generalization of the approach to unknown environments (different lighting, different obstacles) was not addressed in \cite{DECROON2011}. 

In this article, we propose a novel method of detecting obstacles with monocular vision. In particular, it has robots use self-supervised learning to discriminate everything above the horizon line from everything below. Obstacles will appear both above \emph{and} under the horizon line. Therefore, the classification for obstacle pixels should be higher than for objects that only appear below the horizon line (like the ground or floor) or above the horizon line (like the sky or ceiling). In contrast to \cite{ulrich2000appearance}, the robot can learn by itself that certain objects far away do not pose a threat, without actually having to move over them. In contrast to \cite{DECROON2011}, the robot can learn in any environment (indoors and outdoors) by means of self-supervised learning. Interestingly, the proposed self-supervised learning method actually provides a bridge between ground-segmentation and sky-segmentation approaches to obstacle detection, as will be shown by means of image set experiments.

The remainder of the article is organized as follows. In Section \ref{sec:method} the proposed method, its theoretical properties and limitations, and its implementation in this study are explained in more detail. Subsequently, in Section \ref{sec:experiments}, first the setup and then the results of the experiments are discussed. Conclusions are drawn in Section \ref{sec:conclusion}.

\section{Horizon Approach to Monocular Obstacle Detection} \label{sec:method}
\subsection{Explanation of the obstacle detection concept and its limitations}
\begin{figure}
\centering
\includegraphics[width=12cm]{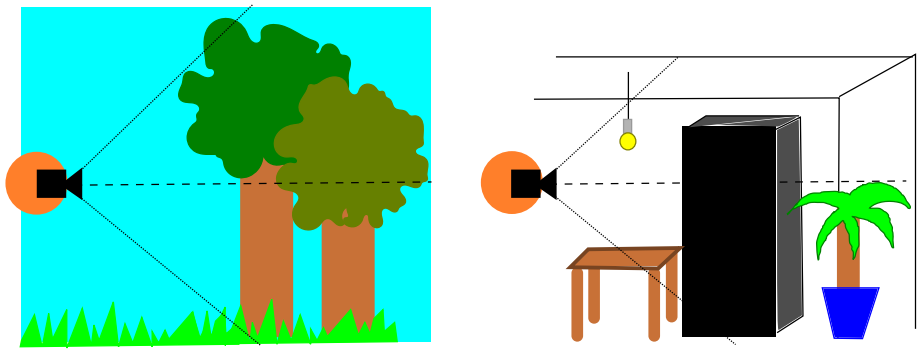} 
\caption{The relation between the observer's eye height (camera) and the horizon line (the dashed line). Obstacles are objects that are present }
\label{fig:concept}
\end{figure}

The proposed obstacle detection method has its foundation in Gibson's ecological approach to visual perception \cite{GIBSON1979}, which is constructed around invariances in an observer's perception. The most well-known visual invariant introduced by Gibson may be optical flow, but in his book Gibson also discusses the horizon line as an invariant (\cite{GIBSON1979}, p. 165): \emph{``The line where the horizon cuts the tree is just as high above the ground as the point of observation, that is, the height of the observer's eye. Hence, everyone can see his own eye-height on the standing objects of the terrain.''}. Figure \ref{fig:concept} shows this relation between the observer's eye height (here shown by a camera icon) and the horizon line (in world coordinates shown by the dashed line). For a point-sized observer that moves at a constant altitude, obstacles indeed intersect with the horizon. Outdoors, the camera will only ever see the sky above the horizon line, as any optical ray under the horizon line will touch a ground-based object (excluding for now reflections of the sky in a mirror-like surface). For indoor environments, if the ceiling has a different appearance than the rest of the environment, it will be identified as a non-obstacle as well. This is actually correct if the robot keeps the same altitude. Conversely, a flat outdoor ground or indoor floor will never surpass the horizon line. The idea behind the proposed method then is to detect obstacles by learning the appearance of objects that intersect with the horizon line.

Figure \ref{fig:concept} illustrates some of the assumptions and properties of this concept. The left image makes clear that an assumption of the method is that the appearance of the obstacle above and below the horizon is similar. In a theoretical case, the camera could be exactly aligned with the transition of one color and texture to another (e.g., on the exact separation point between the green and brown of the tree in our figure). However, given a class of objects of slightly different sizes, and given the natural variation in the height of any flying robot, in practice this is extremely unlikely. The same kind of height variation will make the plant in the right part of the figure (just touching the horizon line) appear as an obstacle as well. The height of the camera is still very important though: what constitutes an obstacle at one height may not constitute an obstacle at another. For instance, the desk in the indoor environment is not an obstacle when flying at the shown camera height, but will be an obstacle when flying considerably lower. Additionally, please note the orange circle around the camera: in theory, the concept is only applicable to a point-sized observer. However, in practice for most flying robots, the body of the agent is small compared to the height variations during learning. Moreover, most ground-based obstacles will significantly extend below and above the horizon. As a consequence, the practical problems for a small-bodied robot are limited. Finally, please note that overhanging obstacles such as the lamp on the right will be detected as an obstacle if they intersect the horizon line.

Outdoors, a flying robot can travel in directions that do not have any obstacle surpassing the horizon line, and rise higher if such a direction is not available (see \cite{DECROON2011}). Indoors the robot cannot do this, as it will be limited by the ceiling. In fact, if truly indoors, the robot is always surrounded by obstacles in all directions. How to determine then an obstacle-free direction in which to travel? For this, multiple solutions exist. In the current article, a flat ground plane is assumed. Together with the height and the image position at which obstacles touch the ground, the distance can then be determined to each obstacle for each image column. This would boil down to the same obstacle detection strategy as in, e.g., \cite{ulrich2000appearance}. However, how the ground is segmented and learned is completely different from that work.

% Other solution: TTC of the limit of the obstacle

\subsection{Implementation}
The advantage of the proposed horizon approach to obstacle detection, is that any robot with an Inertial Measurement Unit (IMU) can continuously learn by itself what objects intersect with the horizon line. Namely, the robot knows its attitude based on the IMU measurements, which in turn means that given a known placement of the camera, the horizon line can be projected in the image. Hence, the robot can employ \emph{self-supervised learning} in which it labels all pixels in the image with respect to being above or below the horizon line. The advantages of self-supervised learning are that: (1) it can happen in the environment in which the robot operates, so that there are only small discrepancies between the training and testing distributions, (2) supervised learning can be fast, and (3) a large amounts of labeled examples will be available to the robot, enabling machine learning methods with great representational complexity such as deep neural networks. 

The key to the approach then is that the robot's interest is not in the classification itself, but in the \emph{uncertainty} with which the classification is performed. Appearance features that appear both above and below the horizon line should have a higher uncertainty than features that only appear on either side. This means that a classifier has to be used that also outputs an (un)certainty value. Many such methods are available, including deep neural networks that can use dropout during test time, e.g., \cite{gal2016dropout}. 

In this preliminary study, we forgo deep neural networks for now, and use a random forest classifier \cite{breiman2001random}. Per pixel a simple feature vector is constructed, concatenating HSV-color values with texture features (1 Local Binary Pattern, LBP, of radius 3 with 24 sampling points \cite{maenpaa2003local}, and the 9 Laws masks \cite{laws1979texture}). The `simplicity' of this implementation actually may be an advantage for both training speed and the computational effort during training and testing - when executed on computationally restricted flying robots.

%===============================================================================

\section{Experiments}
\label{sec:experiments}

\subsection{Experimental Setup}
Three different data sets are studied. The first data set is part of the well-known KITTI data set \cite{Geiger2013IJRR}, which contains videos taken from a driving car. For this study, `drive0001' was used. The second data set is captured in a flight arena. A large part of the arena is a `toy'-like environment, with a green / blue / black artificial grass carpet, and black curtains in the background. However, it does feature differently colored and textured obstacles and mats on the floor. The third data set was made inside an office environment. It is the only data set with a manually labeled floor that can serve as a ground-truth (no pun intended).

For each environment, the data set was split up in a $90\%$ training set and $10\%$ test set. The features were extracted with the Python sci-kit image library and the random forest trained off-line with the sci-kit learn package. We changed the random forest settings, so that 20 trees are used with at least 10 samples per leaf node. The reason for this is to keep tree sizes within bounds. Then, on the test set, the random forest was asked for the probabilities of both labels (above and below the horizon line), and the entropy of the resulting distribution calculated. These values can then be shown per pixel in the image. An ``obstacle map'' is created by thresholding the uncertainty image at the $25^{\mathrm{th}}$ percentile of the uncertainty values experienced in the training set. The obstacle map is also filtered, removing pixels that have fewer than one neighbor detected as `obstacle'. This step is to remove noise from the obstacle map. %  (conserving the image sequence to avoid too big a dependence between training and test set)
% How to threshold?

\begin{figure}[h!]
\centering
\includegraphics[width=12cm]{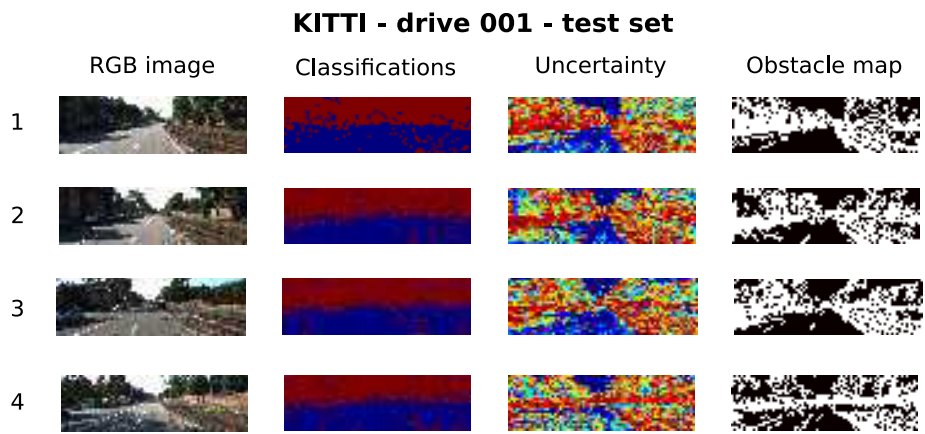}
\caption{Results on the KITTI data set, `drive0001'. From left to right: RGB image, classification results (red is above and blue is below the horizon), uncertainty of the random forest classifications (blue is certain, i.e., an entropy of 0, and red is uncertain, i.e., an entropy of 1), and an obstacle map based on thresholding and filtering the uncertainties (white is obstacle, black is not an obstacle).}
\label{fig:KITTI}
\end{figure}

\subsection{Results} 
Figure \ref{fig:KITTI} shows the results for the KITTI data set. The most striking observation is that the road is clearly segmented as a non-obstacle - this just by learning the difference between objects above and below the horizon. Also the sky is clearly recognized as a non-obstacle. Learning to classify everything below and above the horizon line actually links two known obstacle detection approaches together; while the recognition of the road permits a ground-plane approach to obstacle detection, the recognition of the sky allows a sky-segmentation based approach to obstacle detection. 

The selected images also show that the very simple implementation tested in this study has its limitations; shaded areas lead to higher uncertainties and hence to some spurious obstacle detections. The image in the $4^{\mathrm{th}}$ row has most shade on both the left and right side of the road. This leads to spurious detections, especially on the left side of the road. More complex visual features and / or learners would likely better cope with this type of shading. Finally, despite the sometimes large uncertainties the classification is actually quite accurate (a test performance of $98\%$ correct classifications). Although the classifications are in principle not of interest for the current study, this does imply that one could use them for instance for attitude estimation.

\begin{figure}
\centering
\includegraphics[width=12cm]{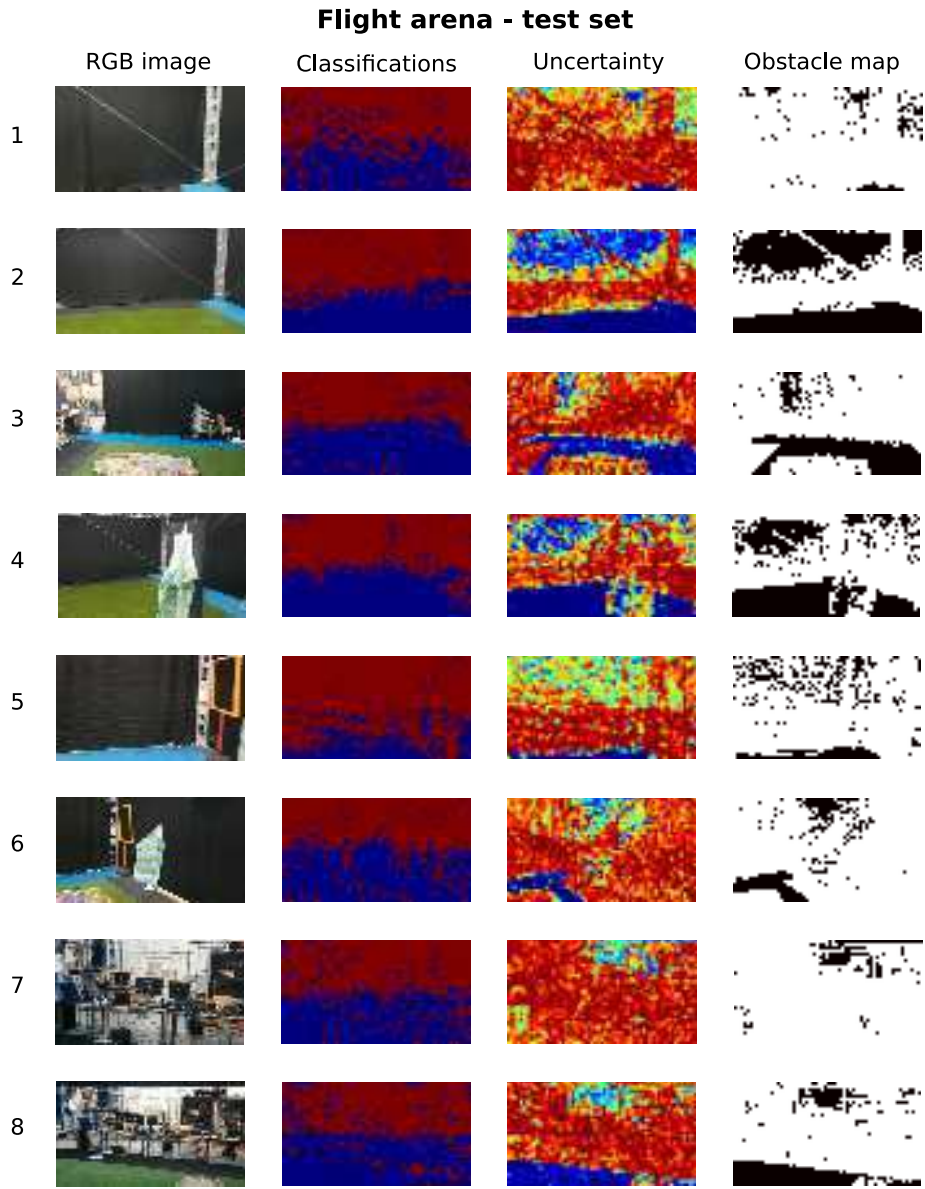}
\caption{Results on the flight arena data set. From left to right: RGB image, classification results (red is above and blue is below the horizon), uncertainty of the random forest classifications (blue is certain, i.e., an entropy of 0, and red is uncertain, i.e., an entropy of 1), and an obstacle map based on thresholding and filtering the uncertainties (white is obstacle, black is not an obstacle).}
\label{fig:Cyberzoo}
\end{figure}

Figure \ref{fig:Cyberzoo} shows the results for the flight arena data set. Again, the most clear ground features are successfully detected. In the flight arena the green and blue artificial grass is successfully indicated as certainly under the horizon line. The classification of the black grass on the ground (visible at the arena's border in rows 1, 2, and 8) has a high uncertainty and is hence mistakenly detected as an obstacle. This is understandable, as the most dominant obstacle type consists of the black curtains around the flight arena. Furthermore, all obstacles in the arena such as the one in row 4 are detected successfully. However, the play mat - with many textures and colors - is also seen as an obstacle. Again, perhaps more complex vision and learning methods would be able to deal with this object. Also, segmenting the obstacle map in row 3 would indicate that the ``mat obstacle'' actually does not touch the horizon line. Row 7 shows an image of the operator zone on the side of the arena. This zone is entirely seen as an obstacle. Indeed, the drone can actually not fly there due to the net. Whether the obstacle detections here are due to the net (present both above and below the horizon line) or due to the similar appearance of the operator zone above and below the horizon line cannot be said with certainty. Row 7 and row 1 also show that obstacles are successfully detected even when they almost fill the entire image. It is in these kinds of situations that an assumption of the area just in front of the robot being free of obstacles is potentially dangerous (as in \cite{lorigo1997visually}). Finally, the classifications are quite good also in this indoor environment, although slightly less so when obstacles fill a large part of the image (see row 1 and 5 for instance).

\begin{figure}
\centering
\includegraphics[width=12cm]{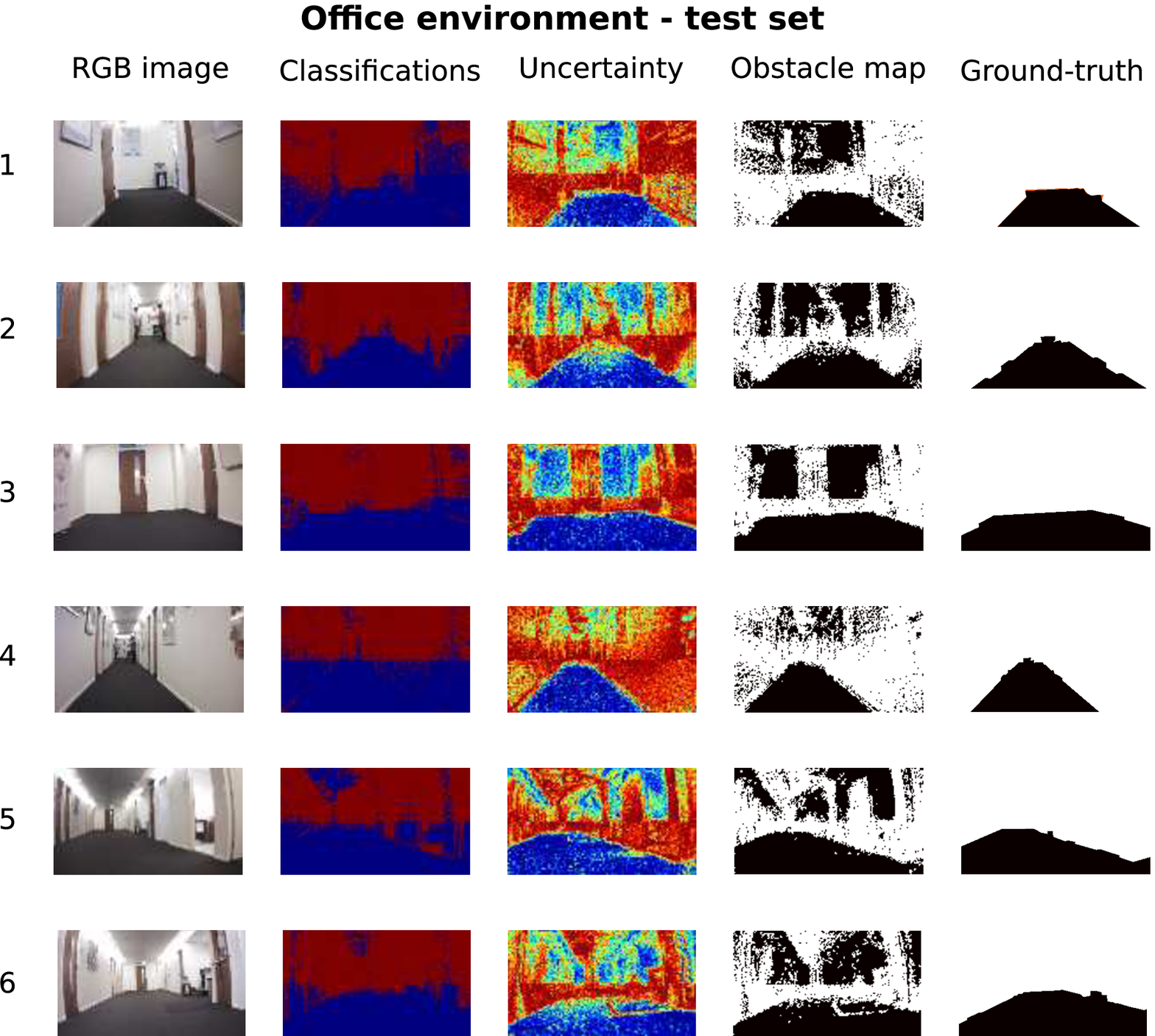}
\caption{Results on the office data set. From left to right: RGB image, classification results (red is above and blue is below the horizon), uncertainty of the random forest classifications (blue is certain, i.e., an entropy of 0, and red is uncertain, i.e., an entropy of 1), an obstacle map based on thresholding and filtering the uncertainties (white is obstacle, black is not an obstacle), and the ground-truth labeling of the floor (black is the floor).}
\label{fig:Marly}
\end{figure}

Finally, the results of the office environment data set are shown in Figure \ref{fig:Marly}. It has to be said that - as many other offices - this environment lends itself well to the proposed method. The appearance does not change a lot over the environment, as can be seen from the 6 rows in the figure, and the floor is generally a uniformly dark blue. This is picked up by the corresponding certainty of the trained classifier. The ceiling, which is in view in multiple images, is not classified as a non-obstacle. The reason for this is likely that its appearance is very similar to that of obstacles such as the walls.

Since for this data set the ground plane has been labeled manually, the performance of the method can be assessed quantitatively. To this end, a Receiver Operator Characteristic (ROC) curve is constructed that captures the true positive ratio (what ratio of obstacle pixels is detected by the thresholded uncertainty) and the false positive ratio (what ratio of floor pixels are mistaken for an obstacle by the thresholded uncertainty). When navigating based on a ground plane assumption, only the obstacle map below the horizon matters. Hence, the ROC curve was constructed by varying the uncertainty threshold for all pixels under the horizon line. Please note that in contrast to the shown obstacle maps in the figures, the ROC curve is made before the spatial noise filtering. 

\begin{figure}
\centering
\includegraphics[width=8cm]{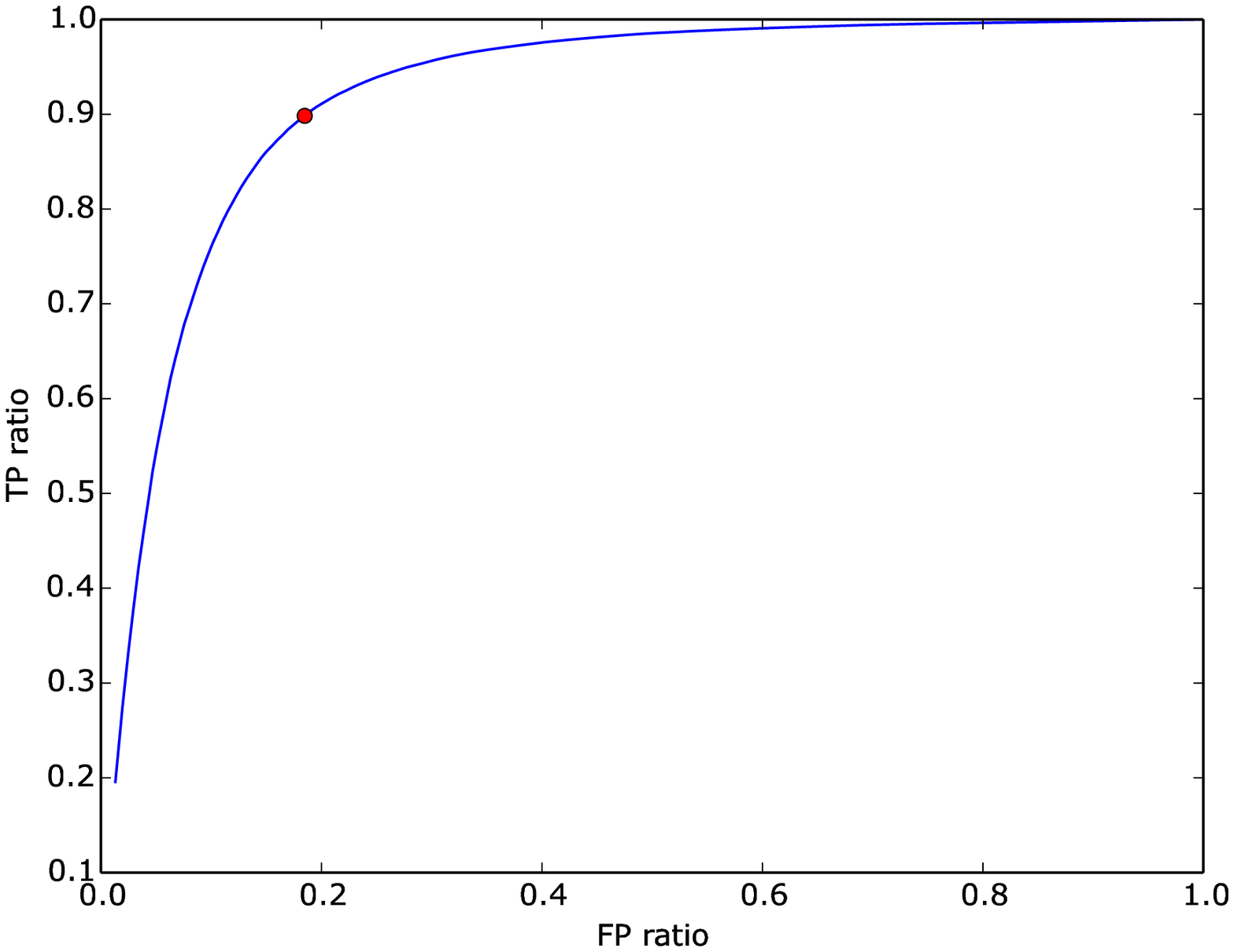}
\caption{ROC curve for all pixels under the horizon, where `positives' are obstacle detections and `negatives' are floor detections. The red dot shows the operating point used for Figure \ref{fig:Marly}}
\label{fig:ROC}
\end{figure}

The resulting ROC curve is shown in Figure \ref{fig:ROC}. It confirms the good performance, with an area under the curve (AUC) of $0.91$. The obstacle maps in Figure \ref{fig:Marly} have been made at the operating point shown by the red dot, with a threshold of $0.64$ on the entropy, a TP ratio of $0.90$, and an FP ratio of $0.18$. Most of the false positives that remain after thresholding are removed by the spatial filtering employed to create the obstacle map. 

%===============================================================================

\section{Conclusion}
\label{sec:conclusion}
In this article, a novel, horizon approach to obstacle detection has been introduced. Robots that know their attitude can learn in a self-supervised way the visual appearance of objects above and below the horizon. The approach is based on the hypothesis that obstacles, defined as objects that intersect with the horizon, will lead to more uncertain classifications than non-obstacles such as a flat ground plane or the sky. A preliminary validation of this hypothesis has been performed with the help of simple visual features, a random forest classifier, and three data sets: (1) the KITTI car data set, (2) an image set taken in a flight arena, and (3) an office environment data set. In the car data set, both the road and sky are successfully detected as non-obstacles. This is interesting, as it suggests that the proposed horizon approach forms a link between ground-plane based obstacle detection and sky segmentation based obstacle detection. In the flight arena and office environments the floor is successfully detected as a non-obstacle. The ceiling is either not visible (in the flight arena) or classified as an obstacle because of its similarity to the walls (in the office). Quantitative evaluation on the labeled data set showed an ROC curve with an AUC of $0.91$, which corresponds to accurate obstacle detection. It has to be emphasized again that this performance was achieved without using the ground labels for training: the robot only attempts to see the difference between everything above and below the horizon line. The results are very encouraging for implementing the method also on board of a drone, which the authors intend to do for the final version of this article (using onboard processing on a Parrot Bebop 2).

\section*{Acknowledgments}
The authors would like to thank Marly Kuijpers, who in the course of her MSc thesis has gathered one of the data sets used in the current study. The proposed method has a patent pending \cite{PATENT_DECROON}.

%% Use plainnat to work nicely with natbib. 

\bibliographystyle{plain}
\bibliography{bibliogdc}

\end{document}